\documentclass{article}
\usepackage{spconf,amsmath,epsfig}

\usepackage{smartdiagram}

\title{Geo-imagery management and statistical processing\\in a regional context using Open Data Cube}
\twoauthors{U.Otamendi, I.Azpiroz, M.Quartulli,I.Olaizola}{Vicomtech Foundation \\ \textit{Basque Research and Technology Alliance (BRTA)} \\ Donostia-San Sebastián, 20009, Spain }{F.J.Perez, D.Alda, X.Garitano}{HAZI Foundation\\\textit{Granja Modelo} \\ Arkaute, 01192, Spain}
\begin{document}
%
\maketitle
\begin{abstract}

We propose a methodology to manage and process remote sensing and geo-imagery data for non-expert users. The proposed system provides automated data ingestion and manipulation capability for analytical data-driven purposes. In this paper, we describe the technological basis of the proposed method in addition to describing the tool architecture, the inherent data flow, and its operation in a specific use case to provide statistical summaries of Sentinel-2 regions of interest corresponding to the cultivation polygonal areas located in the Basque Country (ES).
\end{abstract}
\section{Introduction}
\label{sec:intro}
The use of high spatial resolution optical imagery for land use and land cover change mapping \cite{snyder2019comparison,poyatos2003land,thanh2018comparison} has generated a high demand for efficient geo-imagery storage, processing, and management infrastructures \cite{killough2018overview}. In response to that request, The Committee on Earth Observation Satellites (CEOS) has founded the Open Data Cube (ODC) initiative \cite{ross2017open}, publishing a free and open geospatial exploitation tool \cite{maso2019portal}. Large Spatio-temporal Earth Observation (EO) data volumes are rapidly processed through metadata indexing and ingestion procedures, providing an efficient tool to query remote sensing data.
The main drawback of implementing an ODC-based environment for geo-imagery management and analytics is that it requires an investment in hardware, as well as an initial effort on configuring the system with the considered metadata and product descriptions. Several contributions \cite{maso2019portal} address data governance issues in terms of the use of cloud environment tools such as Google Earth Engine (GEE) \cite{gorelick2017google}. GEE provides a cloud environment, where the analysis of georeferenced data (Earth observation satellites, weather, and climate data) is possible with limited data management effort. This has resulted in an efficient and widely used tool for tasks that range from querying and synchronizing climate reanalysis datasets to the exploitation of georeferenced measurements for e.g. land use analysis \cite{zurqani2018geospatial} or to improve existent plant phenology models \cite{oses2020analysis}. A possible alternative to GEE lies in initiatives such as ODC, which allow local institutions to undertake geo-imagery data management and analysis directly.

The ability to rapidly generate local statistics for a time series of remote sensing images represents a valuable asset for geo-imagery exploitation for rapid mapping.
The possibility to routinely perform this generation in a completely automated manner and in terms that are familiar to a domain expert such as a forester or an agronomist represents an added value point.
In particular, the capability to translate the measurements available in remote sensing image products in the format of a multi-resolution tile pyramid allows the information to be queried by simply specifying a spatial region and a temporal interval of interest and a set of collections to consider as sources.
The possibility for the domain expert analyst to deal with pre-processed, pre-organized, and pre-tiled information content, allows them to focus on the intended application without having to cope with specificities of the original data that stem from the way those were collected.

In this sense, the main goal of the current contribution is to describe and introduce a geo-imagery data management, processing, and exploitation service. This service integrates a methodology intended to provide statistical summaries of regions of interest corresponding to geo-polygonal data. The service is oriented to geospatial data analysts with limited knowledge of remote sensing technology.

In what follows, we present in detail the problem to solve, focusing on the processing of geo-polygons in ODC for the extraction of geo-spatial statistics. We describe the implemented architecture, detailing its main components. We proceed by presenting the imagery characteristics for static and time-dependent products. In addition, we have verified the performance of the service using Sentinel-2 imagery to analyze an approximated quantity of 20000 areas of interest located in the Basque Country (ES).

\section{Proposed Architecture}
\label{sec:architecture}

\begin{figure}
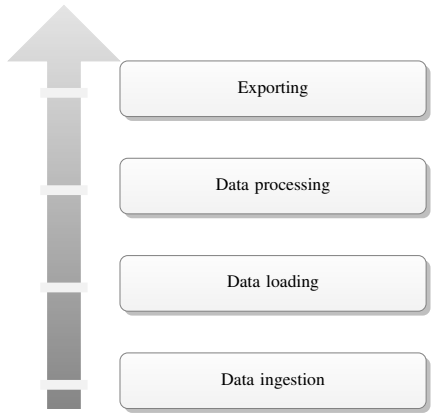

    \begin{center}
    \resizebox{0.7\linewidth}{!}{
        \smartdiagramset{
            priority arrow width=2cm,
            priority arrow height advance=2.0cm,
            priority arrow head extend=0.3cm,
            uniform color list=gray!10 for 5 items}
        \smartdiagram[priority descriptive diagram]{
            Data ingestion,
            Data loading,
            Data processing,
            Exporting
            }
            }
    \end{center}
  \label{fig:global_workflow}
  \caption{
    Intended workflow.  
    The overview of the methodology for Geospatial Data Management and Analysis, from the data ingestion to the data loading and processing for Machine Learning pipelines.}
\end{figure}

In this work, we propose a methodology for Geospatial Data Management and Analysis (see figure 1). This methodology progresses from the insertion of metadata in a geospatial database to the loading of the raster in a data frame including attributes such as the temporal and geographic localization of the specific measurement, to the processing of the obtained data frame --- possibly in terms of Machine Learning pipelines --- to the export of the final results in standard formats usable as input in further analysis.

\begin{figure}[htb]

\begin{minipage}[b]{1\linewidth}
  \centering
 \centerline{\epsfig{figure=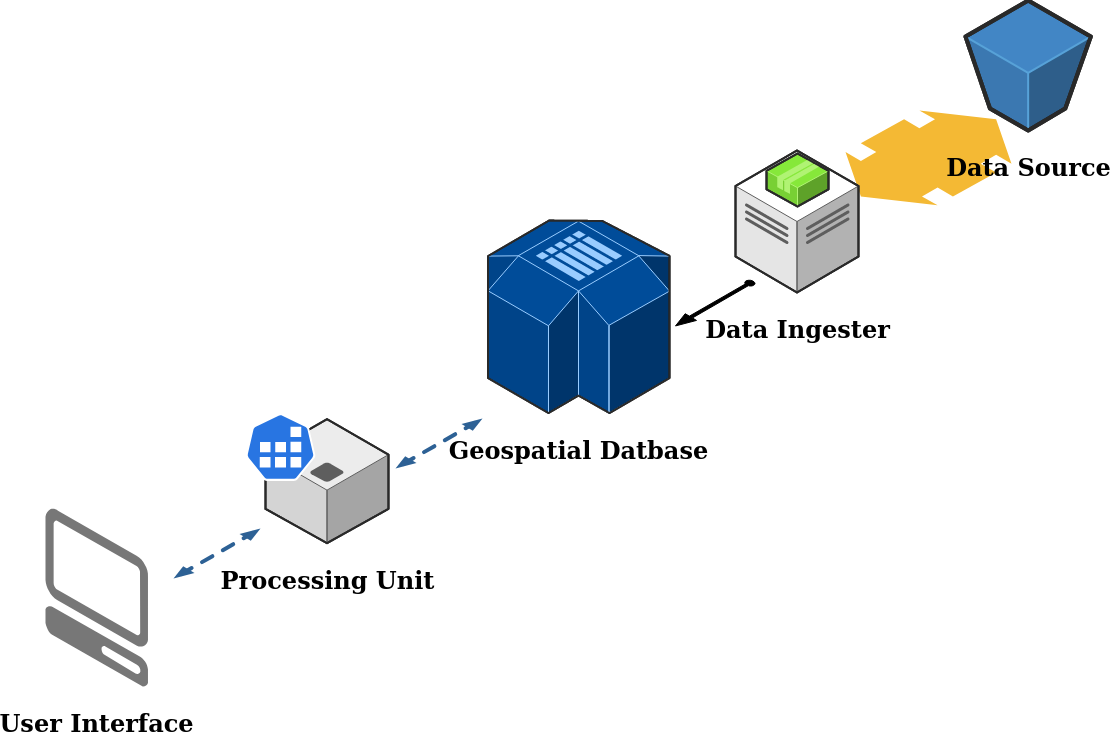 ,width=0.8\textwidth}}

\end{minipage}

\caption{Visual representation of the architecture and the interaction between components. The processing unit is shown as a unique component with a blue icon which represents it is compounded by multiple Docker containers. }
\label{fig:architecture}
\end{figure}

This methodology is integrated into a scalable and robust architecture based on Open Data Cube (ODC) \cite{ross2017open}. This architecture can be used on-premises or deployed as a complete web service. The architecture is built on networked Docker containers \cite{rad2017introduction}. Docker's modularity facilitates management and deployment, both remotely and on-premises. The architecture is divided in three principal components: \emph{Data Sources}, \emph{Data Storage} and \emph{Processing Unit} (see figure \ref{fig:architecture}). 

\emph{Data Sources} are the elements that contain the required information for the analysis. This data requires previous processing and cleaning before the storage of the database. In this work, we propose a processing unit to perform this data ingestion. This unit supports the use of multiple data sources, both on-premises and in the cloud.

The \emph{Data Storage} is the component where the Geospatial data will be stored. After performing the data ingestion from the data sources, this information is stored here. Since it is required by ODC, we use PostgresSQL with the PostGIS extension as the architecture's database. In order to enhance the manageability and operability of the database, we have decided to integrate a Docker containing the pgAdmin tool.

The \emph{Processing Unit} is the core component of the architecture. This unit is responsible to orchestrate and perform the tasks to be performed by the architecture. As mentioned above, this core component relies on the Open Data Cube ecosystem, which offers high processing power. Despite the potential of ODC, it has certain limitations for automated data ingesting and may not be user-friendly for a non-expert user. Therefore, in addition to architecture, we propose a Python library based on ODC to facilitate the analysis.

\subsection{Processing Library}
\label{sec:library}

The processing library is an extension of the ODC with a few improvements that reduce the learning curve for a non-specialized user. The main goal of this library is to provide the user the necessary tools to take advantage of this Geospatial Data Management and Analysis framework with the least possible effort.

Through this library, the user will be able to realize all the necessary tasks to perform the data analysis. These tasks are identified as data ingesting, loading, visualizing, and exporting. 

\textit{Data ingestion} is the first task that needs to be performed before starting the analysis and provided the user with the ability to manage and handle geospatial data in an automated way. As mentioned above ODC has certain limitations to automate the data ingestion and makes difficult the usage of the tool. It requires the geospatial data to be ingested manually, specifying the features and characteristics of each raster image. This, apart from slowing down the process, forces the user to be familiar with the characteristics of the images being processed.

Therefore, we propose a methodology to leverage data diversity by generating metadata descriptions of geo-imagery sources. Specifically, we use data-driven methods to extract metadata and information about e.g. the acquisition geometry, sensor configuration, \textit{et cetera} from a sample image of the product. This process infers information such as the description of telemetry bands, coordinate reference system, resolution, \textit{et cetera}, from the raster products.
Hence, we provide the user with the ability to add more types of geospatial imagery without requiring any prior knowledge of the resource. 

Furthermore, the unit is able to process data periodically from the desired satellite imagery providers, regardless of being on-premises or in the cloud. Thus, this offers the user the possibility to connect the framework to Open Geospatial Consortium (\textbf{OGC}) services (\textit{WCS, WMS, WMTS, et cetera}) or \textit{Amazon Web Service Buckets} such as Sentinel-L2. Therefore, this will provide access to the latest satellite imagery or customized geospatial data files.

\textit{Data loading} allows the user to explore and extract data from the previously ingested data. This tool interacts directly with the database and creates specific queries to load the required data. The user can create customized queries in order to load data relying on the desired location, satellite type, time series, band type, \textit{et cetera}. 
In contrast to the original ODC loading method, in our library, the user can make more complex queries to be able to provide a more detailed analysis. For example, the user will be able to request the data from several types of images (e.g., Sentinel-L1 and Sentinel-L2) at the same time. 

However, the main improvement we have made on the data loading is the processing of polygons. The data loader use previously ingested or manually defined Geo-Polygons to request data from the database. Unlike ODC, when realizing queries relying on Geo-Polygons, our library only retrieves the data of the points inside the polygon. 

Finally, this process provides a data structure called \textit{geopandas}, which contains the loaded data. Geopandas structures are simple to manage and perform analytical operations.

\textit{Data visualization} is a tool that allows the user to view the loaded data simply. The visualization tool uses a plot representation or an interactive map. Using plot visualization simplifies the exploration of the evolution in the selected area as a time series. Regarding the interactive map, each geospatial point is rendered in the map in order to determine whether the loading has been performed correctly and in the desired area.

To conclude, in the \textit{Data Export} phase, previously loaded and processed data is exported in the desired format such as GeoTiff, GeoPandas, CSV, \textit{et cetera}. The main objective of this procedure is to generate datasets from these data for further analysis.

The data export method is designed to load new data into previously exported data files. In this way, this will avoid the need to re-load data that has already been processed in the past. This tool is powerful to periodically update the datasets performing data loading.

\section{Usage example}
\label{sec:typestyle}

In this section, we describe a practical scenario where we detail the operation of the architecture. Prior to geospatial data processing, we have ingested multiple satellite images using various data resources. This ingestion has been performed using on-premise custom geospatial data and the Sentinel-2 registry of open data on \textit{Amazon Web Service}. Subsequently, we have ingested a GeoPackage file containing several Geo-polygons to use in the data loading and analysis process.

\begin{figure}[htb]

\begin{minipage}[b]{.40\linewidth}
  \centering
 \centerline{\epsfig{figure=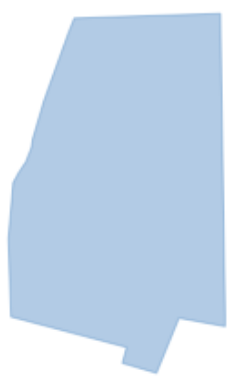 ,width=2.5cm}}

  \centerline{(a) }\medskip
\end{minipage}
\hfill
\begin{minipage}[b]{0.40\linewidth}
  \centering
 \centerline{\epsfig{figure=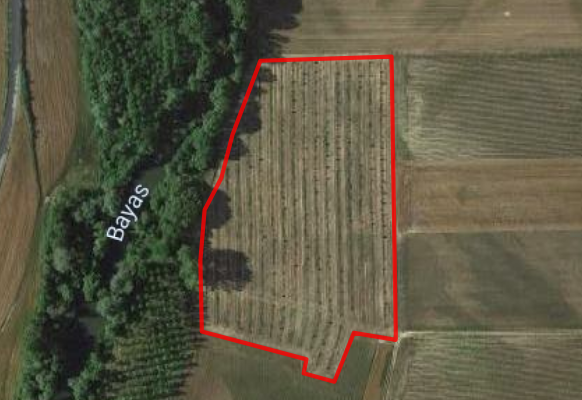,width=5.0cm}}
  \centerline{(b)}\medskip
\end{minipage}
\caption{Geo-polygon visualization example using our proposed python library. (a) Geo-polygon's shape is represented in a plot. (b) Geospatial location of the Geo-polygon on the street view map.}
\label{fig:polygon}
\end{figure}

For simplicity, we have selected a unique polygonal parcel to expose the methodology results (see figure \ref{fig:polygon}). 
The system uses the metadata stored in the geospatial index database to convert the perimeter of the polygon of interest in image pixel coordinates. These coordinates can be used to access the measurement results as performed by the remote sensing system.

These results are transferred to a GeoPandas data frame that can be accessed and processed without having to consider or rely on any of the characteristics of the remote sensing system, which enables the information to be efficiently exploited by application domain experts such as agronomists and foresters without the need for a specific remote sensing background.

Standard data processing procedures are automatically launched on newly available images, for instance, to generate products such as Normalized Difference Vegetation Indices. Time series of measurements (i.e., NDVI) can subsequently be composed and accessed by analytical pipelines.

\begin{figure}[htb]

\begin{minipage}[b]{0.8\linewidth}
  \centering
 \centerline{\epsfig{figure=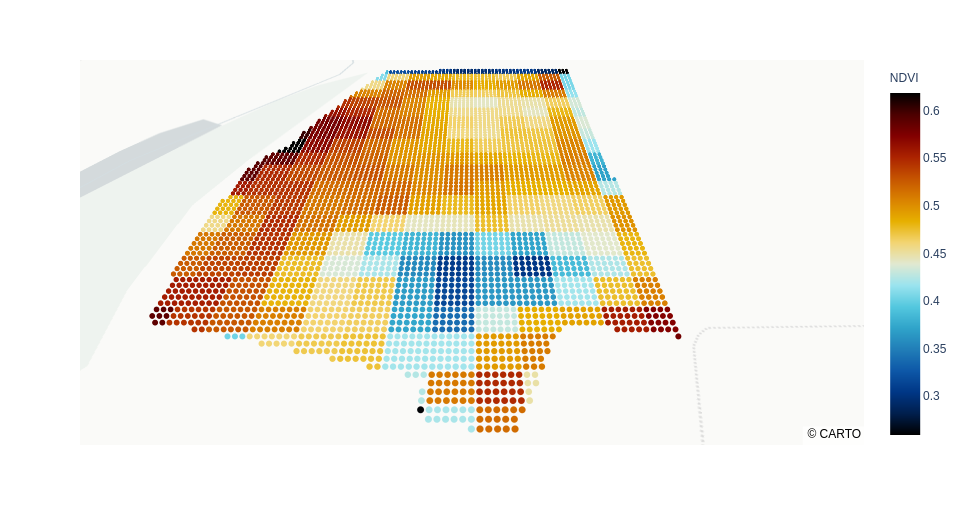,width=0.8\textwidth}}
  \centerline{(a) Result NDVI}\medskip
\end{minipage}
\begin{minipage}[b]{0.8\linewidth}
  \centering
 \centerline{\epsfig{figure=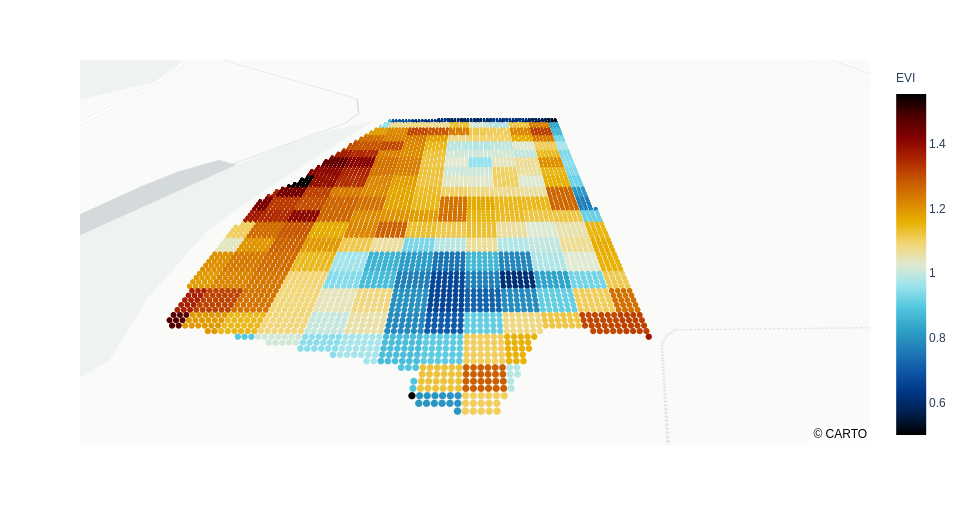,width=0.8\textwidth}}
  \centerline{(b) Result EVI}\medskip
\end{minipage}
\caption{The representation of the polygon of interest can be exposed in a web-based user interface view that depicts it in 3D (a). Each point represents the NDVI value of a 20x20 m land plot. The same representation can be re-used for different indices such as the EVI in (b).}
\label{fig:res}
\end{figure}

The final datasets can be projected to an interactive map visualization that allows users to provide input in the form of interactive selections. This functionality can, for instance, be exploited to enrich the original collection of pre-registered land plots or to provide expert training to machine learning tools (see figure \ref{fig:res}).

\section{Conclusions}
\label{sec:foot}

The extended land usage analysis service solves data storage or management limitations the old local data processing systems have. In contrast to \textbf{vanilla} Open Data cube or Web Coverage Services (\textbf{WCS}), we have introduced a system to provide non-expert users with the ability to manage geospatial data in data-driven algorithms without requiring knowledge of remote sensing or geo-imagery exploitation.

In addition, the automated data ingestion facilitates successful and efficient updates in the data history as well as the subsequent pre-processing procedure. Moreover, the library facilitates the manipulation of those collected measurements for different analytical purposes ranging from time series analysis to the learning of classification and prediction models.

Furthermore, building the architecture on a \textit{dockerized} infrastructure provides the capacity to integrate it in a container-orchestration system such as Kubernetes. In this way, the workload will be balanced depending on the computational cost of the demands, generating new instances of the service. This will increase the scalability of the architecture and enhance the robustness of the system. Moreover, using \textbf{AWS} or \textbf{WCS} as data sources releases the system from data storage and management loads.




\bibliographystyle{IEEEbib}
\bibliography{references}

\end{document}